\def\year{2021}\relax
\documentclass[letterpaper]{article} 
\usepackage{aaai21}  
\usepackage{times}  
\usepackage{helvet} 
\usepackage{courier}  
\usepackage[hyphens]{url}  
\usepackage{graphicx} 
\usepackage{natbib}  
\usepackage{caption} 
\usepackage{multirow}
\usepackage{booktabs}
\usepackage{amssymb}
\usepackage[switch]{lineno}
	\title{Frame-Subtitle Self-Supervision for Multi-Modal Video Question Answering}
\author{
	Jiong Wang, Zhou Zhao, Weike Jin
}
\affiliations{

    College of Computer Science, Zhejiang University, Hangzhou, China\\
    {liubinggunzu, zhaozhou, weikejin}@zju.edu.cn
}

\begin{document}

\maketitle
\begin{abstract}

Multi-modal video question answering aims to predict correct answer and localize the temporal boundary 
relevant to the question. The temporal annotations of questions improve QA performance and interpretability of recent works, but they are usually empirical and costly.
To avoid the temporal annotations, we devise a weakly supervised question grounding (WSQG) setting, where only QA annotations are used and the relevant temporal boundaries are generated according to the temporal attention scores. 
To substitute the temporal annotations, we transform the correspondence between frames and subtitles to Frame-Subtitle (FS) self-supervision, which helps to optimize the temporal attention scores and hence improve the video-language understanding in VideoQA model.
The extensive experiments on TVQA and TVQA+ datasets demonstrate that the proposed WSQG strategy gets comparable performance on question grounding, and the FS self-supervision helps improve the question answering and grounding performance on both QA-supervision only and full-supervision settings.

Note that this paper was rejected by AAAI 2021 and we submit this version only for possible inspirations. We're not planning to release the codes.
\end{abstract}

\section{Introduction}

Video question answering (VideoQA) task \cite{jang2017tgif, kim17story} has consistently attracted considerable attention in recent years and it is challenging by requiring the understanding and reasoning on video and text modalities. Multi-modal VideoQA \cite{tapaswi2016movieqa, lei2018tvqa, lei2019tvqa} provides video frames and associate subtitles, which requires both visual an language understanding in videos to answer corresponding questions. Multi-modal VideoQA is general because the subtitles or dialogs are accessible from videos with the help of the automatic speech recognition (ASR) system  to convert speech into text \cite{miech19how, sun19video}.

Recent multi-modal VideoQA works \cite{lei2018tvqa, lei2019tvqa} extend VideoQA with spatio-temporal annotations and require intelligent systems to simultaneously retrieve relevant moments and detect referenced visual concepts to answer questions. 
The temporal annotations of questions improve the QA performance and interpretability of recent works \cite{kim2020dense, KimMPKY20}, but these annotations are usually empirical and imprecise. Even though \cite{lei2019tvqa} refined the original imprecise temporal annotations, they are still costly to obtain. In this way,  
a Weakly-Supervised Question Grounding (WSQG) setting with only question-answer annotations used is necessary to avoid costly annotation.
The proposed WSQG strategy generates temporal boundaries from the temporal attention scores in VideoQA model, which is inspired by the recent weakly supervised temporal localization works \cite{nguyen18weak, shou18auto}. Considering the unique properties of question grounding, we additionally devise a proposal selection strategy to get complete temporal proposals.


\begin{figure}[t]
	
	\begin{center}
		\includegraphics[scale=0.42]{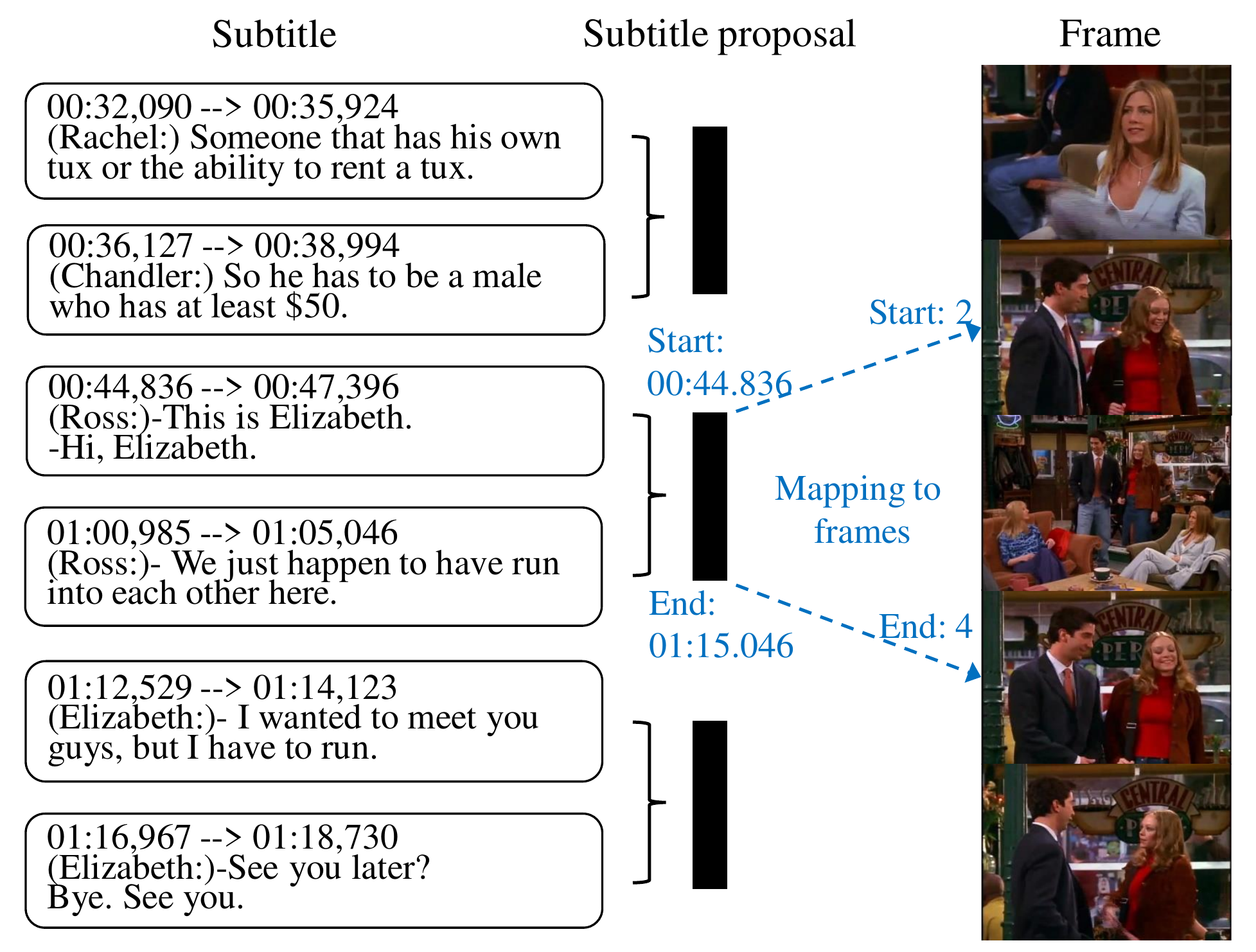}
	\end{center}
	\caption{
		Illustration of the frame-subtitle self-supervision. The adjacent subtitle pairs are connected to subject proposals, which are projected to temporal dimension of frames to get corresponding temporal boundaries.
	}
	\label{fig:motivation}
\end{figure}

 To merge the multi-modal video inputs, most of existing works adopt two-stream structures, which separately encode contextual inputs (subtitles, frames) with QA pairs, then late-fusion with a classifier predicts the final answers. In each stream, \cite{lei2018tvqa} adopts bi-directional LSTM (BiLSTM) to encode both textual and visual sequences, then follow-up work \cite{lei2019tvqa} improves it with convolutional encoder and QA-guided attention. \cite{kim2020dense} employs dense captions to help identify objects and actions in video frames, and hence gives the model useful extra information with explicit textual format to allow easier matching with questions.

We argue that the above works ignore the correspondence between the video frames and subtitles (dialogs), which is adopted as supervision for pre-training video-language joint models  in recent works \cite{miech19how, sun19video}.
In this paper, we devise the frame-subtitle (FS) self-supervision, which grounds video subtitles in video frames. 
As shown in Figure~\ref{fig:motivation}, the video subtitles are provided with time-stamps. We connect adjacent subtitle pairs to subject proposals with start-end timestamps, which are then projected to frames' temporal dimension to get corresponding temporal boundaries. In training phrase, the temporal supervision is used to optimize the temporal attention scores and hence guides video-language understanding in VideoQA model.

In summary, we devise a weakly supervised question grounding (WSQG) strategy and adopt the FS self-supervision to avoid costly temporal question annotations.  Our contributions are threefold. 
\begin{itemize}

	\item 
	We propose a WSQG strategy, where question grounding supervision is not used and the relevant temporal boundaries are generated according to the temporal attention scores. The WSQG strategy can be applied to existing VideoQA models and directly reflects the interpretability. 
	
	\item We transform the correspondence between video frames and subtitles to FS self-supervision for optimizing the temporal attention scores, thus guides video-language understanding in VideoQA model.
	
	\item Extensive experiments on two multi-modal VideoQA datasets show that the proposed WSQG method gets comparable performance on question grounding, and the FS self-supervision consistently improves question answering and grounding performance on QA-supervision only and full-supervision setting, and achieves new state-of-the-art performance.
\end{itemize}

\section{Related Works}

\subsection{Video Question Answering}
Video question answering (VideoQA) is an emerging research field and there are rapid development on datasets and techniques in recent years.  \cite{tapaswi2016movieqa} proposed the MovieQA dataset and extend memory networks to VideoQA domain, following works \cite{na17a, kim2019progressive} devise well-designed memory architectures to enable better interaction on different modalities. 
\cite{jang2017tgif} proposed the TGIF-QA dataset and combine deep appearance and motion features with spatial and temporal attention for accurate question answering. 

Recent multi-modal VideoQA works \cite{lei2018tvqa, lei2019tvqa} extend VideoQA with spatio-temporal annotations and require intelligent systems to simultaneously retrieve relevant moments and detect referenced visual concepts.
 These works prove that question temporal annotations improve accuracy and interpretability of QA models. Following works \cite{kim2020dense, KimMPKY20} use caption as additional input sources and devise modality weighting strategy for better question answering performance.
However, these works ignore the correspondence between frame and subtitles, which can serve as supervision to guide the language and video understanding in VideoQA model.

\subsection{Weakly-Supervised Temporal Action Localization} 
Weakly supervised temporal action localization aims to localize multiple actions in videos with only video-level action labels given. The prevailing methods can be grouped into
two categories: Top-down \cite{wang17untrim, paul18wtalc} and Bottom-up \cite{nguyen18weak, shou18auto} approaches.
 Most of existing bottom-up works follow the Temporal Class Activation Sequence (TCAS) baseline \cite{nguyen18weak}, where the the class-specific attention score is predicted for each frame and consecutively salient attention scores are summarized to temporal boundaries.
Recent works explore backgrounding modeling \cite{lee20back} and generative attention modeling \cite{shi20weak}. 
The proposed WSQG strategy is in a similar way where only QA annotation given and we summarize the temporal attention scores to temporal proposals. 
 Considering the unique properties of question grounding, we devise a proposal selection strategy to avoid the local maximums in attention scores, and hence generates complete temporal proposals. 

\begin{figure*}[t]
	
	\begin{center}
		\includegraphics[scale=0.41]{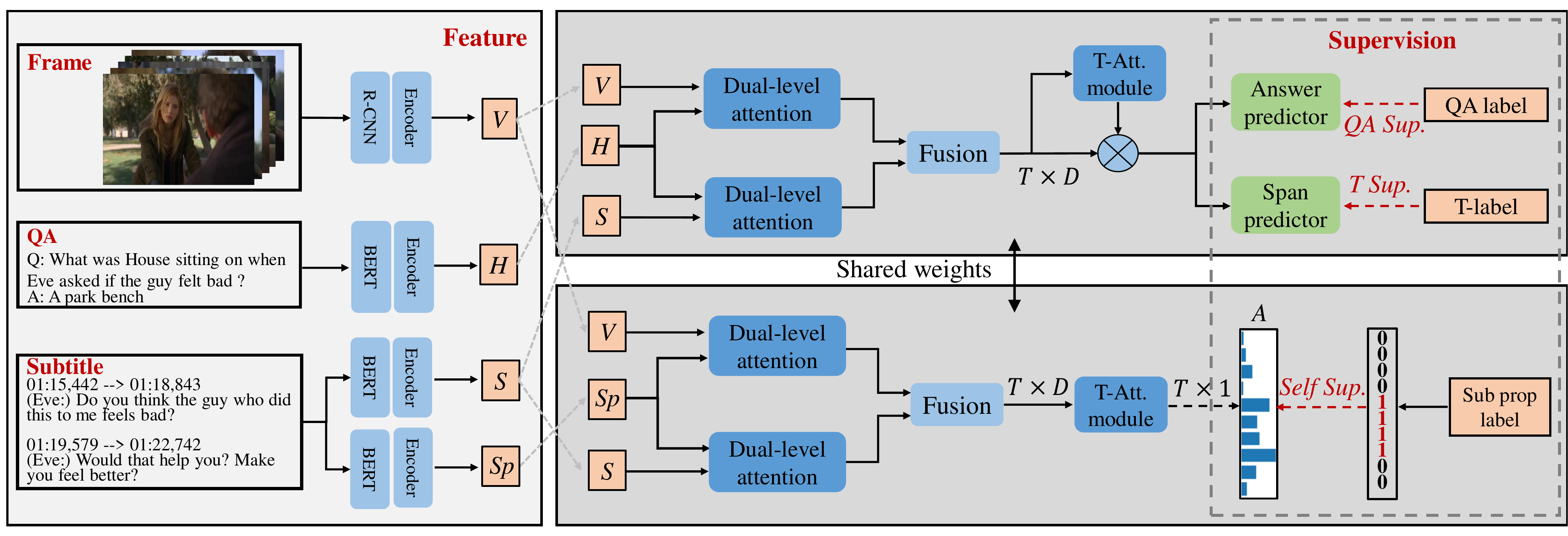}
	\end{center}
	\caption{
		Training pipeline with three kinds of supervision. 
		The video frames, subtitles, and QA pairs are encoded in the left part. The two parts in the right model the multi-modal interaction and achieve question answering and grounding. 
		The QA label denotes correct answer index, the T-label denotes temporal boundary of question, and the Sub prop label is temporal boundary of subtitle proposal.
	}
	\label{fig:pipeline}
\end{figure*}
\subsection{Visual-Text Self-Supervision}  
Recently, there emerge many works \cite{lu19vilbert, li19visual, su20vl} that extend Bert \cite{devlin19bert} in vision-language field as a powerful pre-trained model. 
These works adopt the image-caption self-supervision from the Conceptual Captions \cite{sharma18concept}  or COCO captions \cite{chen15coco} dataset.
As a representative of these works, ViLBERT \cite{lu19vilbert} is pretrained on the Conceptual Caption dataset with masked multi-modal reconstruction and multi-modal alignment prediction. 

In video domain, recent works resort to the frame-dialog self-supervision in narrated instructional videos, where the spoken words are more likely to refer to video content. HowTo100M \cite{miech19how} learns powerful video-language embedding with 136 million video clip-transcribed narration pairs. VideoBERT and ActBERT \cite{sun19video, zhu20act} build on the BERT model to learn bidirectional joint distributions over sequences of visual and linguistic tokens.
These works pre-train powerful video-language joint models by matching video contents to corresponding dialogs, and show excellent generalization capability on downstream video-language tasks. 

In this paper, we transform the correspondence between video frames and subtitles to the frame-subtitle self-supervision, which temporally grounds the subtitles to video frames. 
Different from the above works adopt visual-text self-supervision for pre-training, the proposed video FS self-supervision is adopted to complement the QA supervision for accurate question answering and grounding. 

\section{Proposed Method}

\subsection{Preliminary}
In multi-modal VideoQA, an input instance includes a question $q$ with 5 candidate answers $\left \{ a_{k} \right \}_{k=1}^{5}$, video frames and associated subtitles with timestamps. The goal is to predict correct answer and ground the QA pair temporally. The question and answers are concatenated to 5 QA pairs (hypothesis) $h_k = [q, a_k]$.

For each frame, Faster R-CNN \cite{ren17faster} pre-trained on Visual Genome \cite{kris17genome} is used to detect objects
and extract their region embeddings as the video frame features. The top-20 object proposals are preserved and dimension-reduced from 2048 to 300 by PCA. 
BERT word embeddings with 768 dimensions are used to embed QA pairs and subtitle sentences. 
For each frame, we pair it with two neighboring subtitle sentences based on the subtitle timestamps.
In this way, the resulted aligned subtitles 
are convenient for late fusion with video frame features. We pair adjacent subtitles to subtitle proposals as illustrated in Figure~\ref{fig:motivation}.

Both the object-level embeddings and the word-level embeddings are then projected into a 128-dimensional space using a linear layer with ReLU activation. We adopt convolutional encoder with positional encoding to get the encoded frame, subtitle and QA features with shape preserving following \cite{lei2019tvqa, kim2020dense}.
We denote the number of words in each hypothesis as $L_h$, the aligned subtitles as $L_s$ and the subtitle proposals as $L_{sp}$ respectively. We use $N_o$ as the number of object regions in a frame and $T$, $T_{sp}$ as the frame number, subtitle proposal number respectively. 
Thus we get encoded QA features  $H \in \mathbb{R}^{5 \times L_h \times 128}$, video frame features $V \in \mathbb{R}^{T \times N_o \times 128}$, subtitle features $S \in \mathbb{R}^{T \times L_s \times 128}$ and subtitle proposal features $Sp \in \mathbb{R}^{T_{sp} \times L_{sp} \times 128}$.

\subsection{Model}
The pipeline of our model is inspired by recent multi-modal VideoQA works \cite{lei2019tvqa, kim2020dense} with well-designed cross-attention mechanism. 
We take one QA pair as an example for clarity as shown in Figure~\ref{fig:pipeline}. 
The upper part illustrates the baseline method where the two-stream architecture with late fusion is adopted for question answering and grounding, which is supervised by the QA labels and question temporal labels (T-label). 

In the bottom part, the subtitles are transformed to subtitle proposals with corresponding temporal labels (Sub prop label), which optimize the attention scores in temporal attention module. These two parts share same weights which are optimized by three kinds of supervision.

Concretely in each stream of the upper part, the inputs are query (QA or subtitle proposal) and context (video and subtitle) features. The dual-level attention module is adopted to capture the bi-directional feature interaction on word-level and frame-level.

\textbf{Word-level attention.}  
Taking one encoded video frame feature $v \in \mathbb{R}^{N_o \times 128}$ and one QA pair feature $h \in \mathbb{R}^{L_h \times 128}$ as an example, we apply word-level attention where a similarity matrix is first calculated for all word-region pairs as
\begin{equation}
	Sim_{v} = hv^T,   Sim_{v} \in \mathbb{R}^{L_h \times N_o}.
\end{equation}

  Then $Sim_v$ and $Sim_v^T$ are respectively multiplied on $v$ and $h$, and max pooling is applied on word or region-level to reduce dimension as follow. 
 \begin{equation}
 v_{att} = max(Sim_v v), v_{att} \in \mathbb{R}^{128}.
  \end{equation}
  \begin{equation}
  h_{att} = max(Sim_v^T h), h_{att} \in \mathbb{R}^{128}.
  \end{equation}
  The resulted attentive features are fused together to get the attentive frame feature $v_f \in \mathbb{R}^{128}$ as follow.
  \begin{equation}
  v_f = ([v_{att}; h_{att}; v_{att} \odot h_{att}; v_{att} + h_{att}])W_1 + b_1,
  \end{equation}
  where $W_1$ and $b_1$ are parameters to reduce the 512-d fused features to 128-d. The whole attentive frame features are $V_f \in \mathbb{R}^{T \times 128}$. In similar way, we get the attentive subtitle features $S_f \in \mathbb{R}^{T \times 128}$ in another stream.

\textbf{Frame-level attention.}  
After the word-level attention, we get the attentive frame feature $V_f$ and subtitle features $S_f$. Similar to word-level attention, a similarity matrix $Sim_f \in \mathbb{R}^{T \times T}$ is calculated on frame-level. 
\begin{equation}
Sim_{f} = S_f V_f^T,   Sim_{f} \in \mathbb{R}^{T \times T}.
\end{equation}
Then $Sim_{f}$  and $Sim_{f}^T$ are separately multiplied on $S_f$ and $V_f$ as follow.
\begin{equation}
V_{fatt} = Sim_f^T V_f, V_{fatt} \in \mathbb{R}^{T \times 128}.
\end{equation}
\begin{equation}
S_{fatt} = Sim_f S_f, S_{fatt} \in \mathbb{R}^{T \times 128}.
\end{equation}
The resulted frame-wise attentive features are fused to get the fused features $F \in \mathbb{R}^{T \times 128}$ as follow. 
 \begin{equation}
F = ([V_{fatt}; S_{fatt}; V_{fatt} \odot S_{fatt}; V_{fatt} + S_{fatt}])W_2 + b_2,
\end{equation}
 where $W_2$ and $b_2$ are parameters for dimension reduction to 128-d.
 
\textbf{Temporal attention module.}
Getting the fused features $F$, a temporal attention (T-Att) module, consisting of a fully-connected layer and sigmoid function, is applied to get the attention scores $A \in \mathbb{R}^{T}$  which identify which frames are relevant to questions. Then we get the attentive fused features by multiplying $A$ on $F$. Note that these attention scores are used to generate temporal proposals in the proposed WSQG strategy when testing.

Finally, the attentive fused features are feed to answer predictor and span predictor to get the answer score and temporal boundary, respectively. Concretely, the span predictor predicts the start and end probabilities on temporal dimension, and is optimized by the temporal question supervision as \cite{lei2019tvqa} do. Given the softmax normalized start and
end probabilities $p_{st}$ and $p_{ed}$, we use cross-entropy loss as follow.
\begin{equation}
loss_{span} = -\frac{1}{2}(log p_{st} + log p_{ed}).
\end{equation}
The QA loss is calculated as 
\begin{equation}
	 loss_{qa} = -log(\frac{e^{p_g}}{\sum_{k=1}^{5}e^{p_k}}),
\end{equation}
where $p_g$ is predicted answer score for the ground-truth index.

\subsection{Video Frame-Subtitle Self-Supervision}
As illustrated in Figure~\ref{fig:motivation}, the adjacent subtitles are connected to subtitle proposals. The reason is only the start-times of subtitle are given in the meta-data, so we connect adjacent subtitles to get their rough start-end time-span.

In the bottom part of Figure~\ref{fig:pipeline}, we replace the QA features with subtitle proposals and feed them to dual-level attention, fusion and T-Att modules with contextual features. The predicted temporal scores $A \in \mathbb{R}^{T}$ are directly supervised by the subtitle proposal labels (Sub prop label).

Note that the word-level attention in the bottom part is slightly different from the upper part, the reason is we don't want to introduce more computations on the baseline method. 

Before the dual-level attention, we first sum the word and region-level subtitle, video and subtitle proposal features to frame-level features.
So the word-level attention module  actually capture coarsely frame-level interaction between subtitle proposals and contextual features. The follow-up frame-level attention is same as the upper part. 

We found this modification doesn't cause performance drop when testing, and only slightly increases the computations on baseline part because the computation of word-level attention is much larger than the frame-level counterpart.

The predicted temporal attention scores $A \in \mathbb{R}^{T}$ reflect which frames are relevant to the subtitle proposals. To calculate the loss between the $A$ and Sub prop label, a ranking loss and a Binary Cross-Entropy (BCE) loss are complemented for optimization.

The ranking loss forces the average attention scores inside the temporal boundary ($A_{in}$) to be larger than the outside counterparts ($A_{out}$) as follow. 
\begin{equation}
	loss_{rank} = 1 + avg(A_{out}) -avg(A_{in}).
\end{equation}
The BCE loss enforces the attention scores of frames inside the temporal boundary to be 1 and outside to be 0, which is defined as follow. 
\begin{equation}
loss_{bce} = - \frac{1}{T_{in}}\sum_{i=1}^{T_{in}}(log A_{in}^{i}) - \frac{1}{T_{out}}\sum_{j=1}^{T_{out}}(log (1 - A_{out}^{j})),
\end{equation}
where $T_{in}$ and $T_{out}$ are the number of frames inside and outside the temporal boundary.
The total self-supervision loss is defined as 

\begin{equation}
loss_{self} = loss_{io} + loss_{bce}.
\end{equation}

In this way, the FS self-supervision guides the temporal attention module to generate reasonable attention scores for language queries and 
helps the VideoQA model better understand the video-language interaction. 
In practice, it improves the performance of answer predictor ans span predictor in the upper part, demonstrating it is complementary to other two kinds of supervision.

\subsection{Weakly-Supervised Question Grounding (WSQG)}
\begin{figure}[t]
	
	\begin{center}
		\includegraphics[scale=0.3]{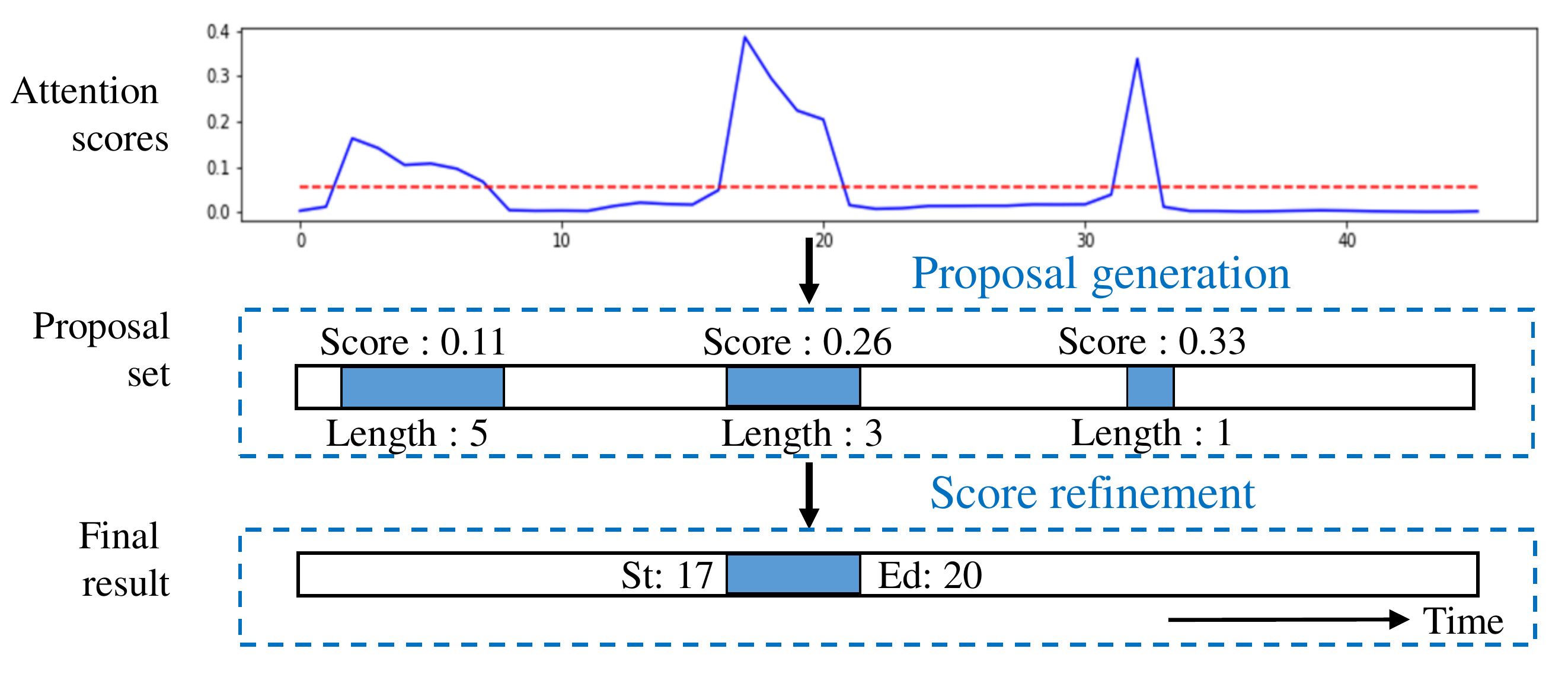}
	\end{center}
	\caption{
		Proposal generation and refinement process from attention scores. The red dashed line denotes threshold.
	}
	\label{fig:proposal}
\end{figure}
To generate consecutive proposals from temporal attention scores, we propose a WSQG strategy which is inspired by the weakly-supervised temporal action localization works \cite{nguyen18weak, shou18auto}. But different from these works which localize multiple actions with multiple proposals, we aim to find the most relevant temporal proposal to the question. So we further devise a proposal selection method, which refines the proposal scores with a length-aware term, to avoid the influence of local maximum in attention scores. 

As shown in Figure~\ref{fig:proposal}, getting the temporal scores $A \in \mathbb{R}^{T}$,  we first set a threshold $A_t$, 
 then consecutive frames whose scores are larger than $A_t$ are connected to temporal proposals which construct a proposal set. Note that we found the attention scores have different magnitudes for different instances, so we set a dynamic threshold which is equal to the average score of $A$. 

For each proposal, we calculate the proposal score, which is the mean attention score inside its time-span as follow:

\begin{equation}
A_P = \sum_{k = st}^{ed}A_{k},
\end{equation} 
We expect the proposal with highest score to be the temporal grounding result. 

In practice we found that the proposal scores will be dominate by some local maximums with quite short temporal boundary, as illustrates by the third proposal  of Figure~\ref{fig:proposal}. So we devise length-aware score refinement strategy where the proposal scores are refined by multiplying a proposal length-aware term as follow:

\begin{equation}
 A_{P_r} = A_P * (ed - st)^{\alpha}, 
\end{equation} 
where $\alpha$ is a parameter to control the effect of proposal length, and we set it to 0.5 in our configuration. Then the proposal with highest refined score is selected as final prediction.

Note that the temporal attention module is a plug-to-play module and is applicable to existing VideoQA models, so the proposed WSQG strategy is suit for existing VideoQA models for question grounding to reflect the interpretability.

\subsection{Four Level of Supervision Settings}
\begin{table}[t]
	\setlength{\abovecaptionskip}{0.cm}
	\setlength{\belowcaptionskip}{-0.cm}
	
	\begin{center}
		\resizebox{.4\textwidth}{!}{
			\begin{tabular}{l | c  c  c}
				\toprule
				Setting   & QA sup. & QG sup. & Self sup.\\
				\hline
				\multicolumn{3}{c}{QA supervision}\\
				QA only & $\surd$ &&\\
				QA + self& $\surd$&&$\surd$\\
				\hline
				\multicolumn{3}{c}{Full supervision}\\
				Full& $\surd$ &$\surd$&\\
				Full + self& $\surd$ &$\surd$&$\surd$\\
				\bottomrule
				
			\end{tabular}
		}
	\end{center}
	\caption{Four levels and three kinds of supervision.}
	\label{tab:supervision}
\end{table}
Together with the proposed FS self-supervision, there are three kinds of supervision for optimizing the VideoQA model.  
As shown in Table~\ref{tab:supervision}, we incrementally summarize four level of supervision settings, including QA supervision only (\textbf{QA only}), QA supervision and self-supervision (\textbf{QA + self}), QA and question grounding (QG) supervision (\textbf{Full}), full supervision and self-supervision (\textbf{Full + self}). 
The QA supervision is adopted without exception and question answering is consistent in training and testing phrase for four settings, while question grounding is not. 
In QA only and QA + self settings, where the temporal supervision is not adopted, the proposed WSQG strategy is used to get temporal proposals from attention scores. In Full and Full + self settings, the temporal proposals are generated from the span predictor in Figure~\ref{fig:pipeline}. 
Specifically in the strongest level, the Full + self setting, three kinds of supervision are adopted and the total training loss is 
\begin{equation}\label{eq:sup}
	loss = loss_{qa} + \lambda_1 loss_{span} + \lambda_2 loss_{self}.
\end{equation}
where $\lambda_1$ and $\lambda_2$  are the hyper-parameters to control the balance of three losses.
\section{Experiments}
\subsection{Datasets}
\textbf{TVQA}. 
TVQA \cite{lei2018tvqa} is a large-scale multi-modal video QA dataset based on 6 popular TV shows spanning 3 genres: medical dramas, sitcoms, and crime shows. It consists of 152.5K QA pairs from 21.8K video clips, spanning over 460 hours of video. The training set consists of 122,039 QAs from 17,435 clips, the validation set 15,252 QAs from 2.179 clips and test set consist of 7,623 QAs from 1,089 clips.
The video contents consist of video frames and subtitles. The multiple-choices form questions-answer pairs are human-annotated and designed to be compositional. 
TVQA dataset requiring systems to jointly comprehend subtitles-based dialogue to correctly answer questions and localize relevant moments within video frames.

\noindent \textbf{TVQA+}. TVQA+ \cite{lei2019tvqa} is a subset of the TVQA dataset and augments TVQA dataset with frame-level bounding box annotations,  which links video objects to visual concepts in questions and answers. %
TVQA+ additionally refine the temporal boundary so the temporal annotations are more precise than TVQA. 
TVQA+ is also the first dataset that combines moment localization, object grounding  and question answering. 
In this paper, we only consider the temporal grounding on TVQA+ dataset.

\subsection{Implementation Details}
The training and testing pipeline on TVQA and TVQA+ datasets follow previous works \cite{lei2018tvqa, lei2019tvqa}. 
The frames are extracted at 0.5 FPS and the questions are grounded in frames' temporal dimension.
RoBERTa \cite{liu19roberta} is used to embed words in QA pairs and subtitle on TVQA \cite{kim2020dense} and BERT word embeddings are used on TVQA+ \cite{lei2019tvqa}.  
The Adam optimizer is used with batch size of 16, initial learning rate is 0.001, which is dropped to 0.0002 after 10 epochs. 
On TVQA+ dataset, the spatial supervision is also adopted for fair comparison with existing works.
The loss weights for different supervision in Equation~\ref{eq:sup} are set to $\lambda_1 = 0.5, \lambda_2 = 0.25$.

\subsection{Evaluation Metric}
The experimental evaluation follows previous works \cite{lei2018tvqa, lei2019tvqa}. The question answering performance is measured by classification accuracy (Acc.). Same as language-guided video moment retrieval works \cite{gao2017tall},  Temporal mean Intersection-over-Union (T. mIoU) is adopted to evaluate question grounding performance. 
We also report the R@1, IoU=m score, 
which measures the percentage of the top-1 predicted temporal boundaries having IoU with ground-truth larger than m.
The Answer-Span joint Accuracy (ASA) is used to jointly evaluate both answer prediction and span prediction, where a prediction is correct only when the predicted span has an IoU $\ge$ 0.5 with the GT span and the answer prediction is correct.

\subsection{Comparison with State-of-the-Arts}
Even though TVQA and TVQA+ datasets require both question answering (QA) and grounding (QG), some recent works only consider question answering. The most comparable works are MSAN \cite{KimMPKY20} and STAGE \cite{lei2019tvqa} which consider both QA and QG. In this subsection, we give comprehensive comparison with these works.
\begin{table}[t]
	\setlength{\abovecaptionskip}{0.cm}
	\setlength{\belowcaptionskip}{-0.cm}
	
	\begin{center}
		\resizebox{.48\textwidth}{!}{
			\begin{tabular}{l| c c c}
				\toprule
				Method & Acc. & T. mIoU & ASA\\
				\hline
				\multicolumn{4}{c}{QA supervision}\\
				{Two-stream \cite{lei2018tvqa}}&67.70&-&-\\
				{PAMN \cite{kim2019progressive}}&66.22&-&-\\
				{Multi-task \cite{KimMKKY19}}&66.38&-&-\\
				QA only*&69.13&29.28&20.78\\
				QA + Self*&\textbf{72.13}&\textbf{33.60}&\textbf{25.71}\\
				\hline
				\multicolumn{4}{c}{Full supervision}\\	
				{STAGE \cite{lei2019tvqa}}&70.50&-&-\\
				{MSAN \cite{KimMPKY20}}&71.13&30&-\\
				{DenseCap (Kim, Tang et al. 2020)}&73.34 & -& -\\
				{Full*}&73.59 & 39.80&29.66\\
				Full + Self*&\textbf{74.20}&\textbf{40.49}&\textbf{30.88}\\
				\bottomrule
				
			\end{tabular}
		}
	\end{center}
\caption{Comparison with state-of-the-art works on TVQA validation set in QA supervision and full supervision setting. * denotes the results are implemented by us.}
	\label{tab:tvqa}
\end{table}

\textbf{TVQA}. In Table \ref{tab:tvqa}, we compare with existing works with different settings on the TVQA dataset. 
It is clear that our ``QA only'' setting outperforms existing counterpart works, the reason is we adopt dual-level attention mechanism and RoBERTa word embeddings following \cite{kim2020dense}.
When paired with the proposed FS self-supervision, the ``QA + Self'' setting gets considerable increments on QG and QG performance, even outperforms MSAN which adopts full supervision. 

In the bottom part, we can see that our full supervision setting slightly outperforms the State-of-the-art works on QA and largely outperforms them on QG task. 
The ``Full + Self'' setting achieves new state-of-the-art performance on QA and QG, demonstrating that the proposed FS self-supervision is complementary with the original question answering and grounding supervision. 
It can also be seen that the increments on QA only setting are more distinct, so we expect the FS self-supervision will be a substitute to the costly temporal annotations, and be intensively explored in the future works.

\textbf{TVQA+}. 
In Table \ref{tab:tvqa+}, we compare with existing works in different settings on the TVQA+ dataset. It can be seen that our QA performance on QA only setting is slightly low, but the QG performance with the proposed WSQG strategy used is comparable to STAGE with full supervision. 
When paired with self-supervision, the ``QA + Self'' setting gets consistent improvement on QA and QG accuracy, surpassing counterpart works. 

In the bottom part, the Full setting largely exceeds STAGE on QG performance. When paired with FS self-supervision, ``Full + Self'' setting gets consistent increments and achieves new state-of-the-art results on three metrics.

\begin{table}[t]
	\setlength{\abovecaptionskip}{0.cm}
	\setlength{\belowcaptionskip}{-0.cm}
	
	\begin{center}
		\resizebox{.48\textwidth}{!}{
			\begin{tabular}{l| c c c}
				\toprule
				Method & Acc. & T.mIoU & ASA\\
				\hline
				\multicolumn{4}{c}{QA supervision}\\
				{Two-stream \cite{lei2018tvqa}}&62.28&-&-\\
				{STAGE \cite{lei2019tvqa}}&68.31&-&-\\
				QA only*&67.88&30.30&21.98\\
				QA + Self*&\textbf{69.47}&\textbf{32.03}&\textbf{23.57}\\
				\hline
				\multicolumn{4}{c}{Full supervision}\\	
				{STAGE \cite{lei2019tvqa}}&72.56&31.67&20.78\\
				{Full*}&72.62 & 39.84&31.36\\
				Full + Self*&\textbf{73.22}&\textbf{40.78}&\textbf{32.05}\\
				\bottomrule
				
			\end{tabular}
		}
	\end{center}
	\caption{Comparison with state-of-the-art works on TVQA+ validation set in QA supervision and full supervision setting. * denotes the results are implemented by us.}
	\label{tab:tvqa+}
\end{table}
\begin{figure*}[t]
	
	\begin{center}
		\includegraphics[scale=0.19]{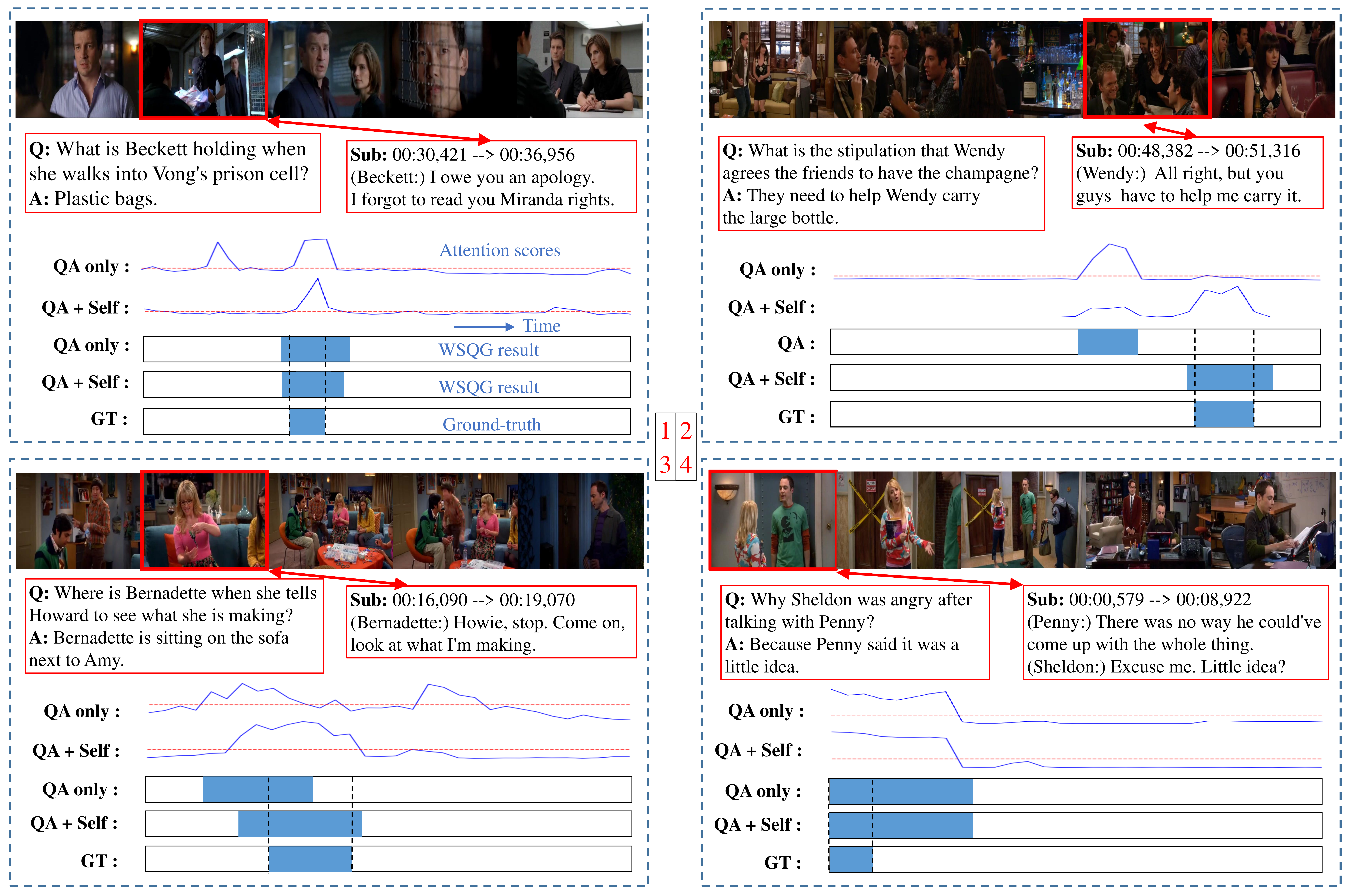}
	\end{center}
	\caption{
		Qualitative results on TVQA val set. The question-answer pairs and relevant subtitles are provided in the red boxes. The frame inside ground-truth (GT) span is highlighted with red box bounded. 
		The attention scores, grounding results of WSQG with QA only and QA + Self supervision settings are provided for comparison with GT.
	}
	\label{fig:quali}
\end{figure*}

\begin{table}[t]
	\setlength{\abovecaptionskip}{0.cm}
	\setlength{\belowcaptionskip}{-0.cm}
	
	\begin{center}
		\resizebox{.48\textwidth}{!}{
			\begin{tabular}{l| c c c c}
				\toprule
				\multirow{2}{2cm}{Method}  & \multicolumn{3}{c}{R@1, IoU =}  & \multirow{2}{1cm}{T.mIoU} \\
				 &0.3 &  0.5 &  0.7 & \\
				\hline
				WSQG w/o refinement &43.26&24.17&9.55&27.69\\
				WSQG (QA only) &45.41&24.40&9.57&29.28\\
				WSQG (QA + Self)  &\textbf{53.26}&\textbf{30.67}&\textbf{11.54}&\textbf{33.60}\\
				\bottomrule
			\end{tabular}
		}
	\end{center}
	\caption{Ablation of WSQG strategy for question grounding with different settings on TVQA val set.}
	\label{tab:grounding}
\end{table}
\begin{table}[t]
	\setlength{\abovecaptionskip}{0.cm}
	\setlength{\belowcaptionskip}{-0.cm}
	
	\begin{center}
		\resizebox{.43\textwidth}{!}{
			\begin{tabular}{l| c c c}
				\toprule
				Scale & Acc. & T.mIoU & ASA\\
				\hline
				Scale (2)&\textbf{72.13}&\textbf{33.60}&\textbf{25.71}\\
				Scale (3)&71.76&33.42&24.75\\
				Scale (4)&71.25&33.26&24.43\\
				Scale (2, 3, 4)&71.78&33.31&24.54\\
				\bottomrule
				
			\end{tabular}
		}
	\end{center}
	\caption{Comparison of different scale of subtitle proposals on TVQA validation set. Scale (3) denotes three adjacent subtitles are connected to one subtitle proposal.}
	\label{tab:scale}
\end{table}

\subsection{Ablation Study}
In this subsection, we give ablation study on the proposed WSQG strategy and the FS self-supervision. 
In Table~\ref{tab:grounding}, the QG performances of WSQG in different settings are reported.
As can be seen,  the proposed proposal score refinement strategy helps WSQG improve the R@1, IoU=m and T.mIoU scores. It can also be seen that the QA + Self setting gets largely increments on the grounding performance, which is benefited from the the FS self-supervision. 
 
 Note that we connect adjacent two subtitles to subtitle proposals, and it is natural to explore proposals with longer scale (more than two subtitles in one proposal) or multi-scale. We give ablation study on different scale of subtitle proposals in Table~\ref{tab:scale}. It can be seen that the adopted Scale (2) gets best performance on question answering and grounding, while longer scale and multi-scale proposals get slightly worse performance.

\subsection{Qualitative Analysis}
 
 Figure~\ref{fig:quali} illustrates qualitative results of the proposed WSQG strategy and the FS self-supervision on TVQA dataset. 
 The proposed WSQG strategy helps to summarize temporal attention scores to temporal proposals, and
 there are  two observations can be summarized.

 First, the attention scores of QA only setting sometimes response on different time-spans, in which some may be falsely matched (Example~1, 3). The scores of QA + Self, by contrast, are more concentrated on the relevant frames. We suppose the reason is training with Self-supervision forces the VideoQA model to focus on the most relevant frames while suppress the irrelevant ones, which benefits both question answering and grounding.
 Second, we found some questions are ambiguous to determine their temporal boundaries, and the annotators may empirically focus on some contents.
 As shown in example 4, given the question, it is not clear enough whether to localize the time-span where ``Sheldon is angry'' or ``Sheldon is talking with Penny''. The VideoQA models tend to localize the temporal spans where ``Sheldon is talking with Penny'' and ``Sheldon is angry'', while the GT only cover the span where ``Penny say it is a little idea''. 
 It reflects that the temporal annotations may be empirical to different annotators who focus on different contents.

\section{Conclusion}
In this paper, we consider question answering and grounding in multi-modal VideoQA. We devise the frame-subtitle self-supervision and a weakly supervised question grounding strategy in multi-modal VideoQA, to alleviate the empirical and costly temporal annotations. The WSQG strategy is applicable to existing VideoQA models to reflect the interpretability. The FS self-supervision helpS to regular the temporal attention scores, thus guide video-language understanding in VideoQA models. It improves question answering and grounding performance in both QA only and full supervision setting, and achieves new state-of-the-art results.

\small
\bibliography{lvqa}
\bibliographystyle{aaai21}

\end{document}